\documentclass[conference]{IEEEtran}
\IEEEoverridecommandlockouts
\usepackage{cite}
\usepackage{amsmath,amssymb,amsfonts}
\usepackage{graphicx}
\usepackage{textcomp}
\usepackage{xcolor}
\def\BibTeX{{\rm B\kern-.05em{\sc i\kern-.025em b}\kern-.08em
    T\kern-.1667em\lower.7ex\hbox{E}\kern-.125emX}}

\usepackage{amssymb}
\usepackage{amsmath}
\usepackage{commath}
\usepackage{mathtools}
\usepackage{subcaption}
\usepackage[linesnumbered,ruled]{algorithm2e}
\usepackage{algpseudocode}
\usepackage{array}
\usepackage{booktabs}
\begin{document}

\title{Deep Variational Clustering Framework for \\
Self-labeling of Large-scale Medical Images}

\author{Farzin Soleymani\textsuperscript{\rm 1}, Mohammad Eslami\textsuperscript{\rm 2}, Tobias Elze\textsuperscript{\rm 2}, Bernd Bischl\textsuperscript{\rm 3}, Mina Rezaei\textsuperscript{\rm 3} \\
\\
\textsuperscript{\rm 1}{Department of Electrical and Computer Engineering, Technical University of Munich, Germany}\\
\textsuperscript{\rm 2}{Massachusetts Eye and Ear Hospital, Harvard Medical School, Boston, MA, USA}\\
\textsuperscript{\rm 3}{Department of Statistics, LMU Munich, Germany} \\
mina.rezaei@stat.uni-muenchen.de}

\maketitle

\begin{abstract}\label{abstract}

One of the most promising approaches for unsupervised learning is combining deep representation learning and deep clustering. Recent studies propose to simultaneously learn representation using deep neural networks and perform clustering by defining a clustering loss on top of embedded features. Unsupervised image clustering naturally requires good feature representations to capture the distribution of the data and subsequently differentiate data points from one another. Among existing deep learning models, the generative variational autoencoder explicitly learns data generating distribution in a latent space. We propose a Deep Variational Clustering (DVC) framework for unsupervised representation learning and clustering of large-scale medical images. DVC simultaneously learns the multivariate Gaussian posterior through the probabilistic convolutional encoder, and the likelihood distribution with the probabilistic convolutional decoder; and optimizes cluster labels assignment. Here, the learned multivariate Gaussian posterior captures the latent distribution of a large set of unlabeled images. Then, we perform unsupervised clustering on top of the variational latent space using a clustering loss. In this approach, the probabilistic decoder helps to prevent the distortion of data points in the latent space, and to preserve local structure of data generating distribution. The training process can be considered as a self-training process to refine the latent space and simultaneously optimizing cluster assignments iteratively. We evaluated our proposed framework on three public datasets that represented different medical imaging modalities. Our experimental results show that our proposed framework generalizes better across different datasets. It achieves compelling results on several medical imaging benchmarks. Thus, our approach offers potential advantages over conventional deep unsupervised learning in real-world applications. The source code of the method and of all the experiments are available publicly at: https://github.com/csfarzin/DVC

\end{abstract}

\section{Introduction} \label{introduction}

Deep learning algorithms have made outstanding results in many domains such as computer vision, natural language processing, recommendation systems, and medical image analysis. However, the outcome of current methods depends on a huge amount of training labeled data, and in many real-world problems such as medical image analysis and autonomous driving, it is not possible to create such an amount of training data. Learning from unlabeled data can reduce the deployment cost of deep learning algorithms where it requires annotations from experts such as medical professionals and doctors. 

Clustering is a fundamental and challenging task of unsupervised learning that aims to group similar data points together without supervision. Unsupervised cluster algorithms were researched widely in terms of density-based modeling, centroid-based modeling, self-organization maps, and grouping algorithms.
In recent years, several approaches have performed image clustering on top of features extracted by a deep neural network (DNN)~\cite{caron2018deep,banerjee2005clustering,rezaei2021learning}. Learning deep representation from data helps to improve cluster analysis compared to traditional centroid-based clustering such as K-means~\cite{arthur2006k}. Deep embedded clustering (DEC) methods train an autoencoder to map a high-dimensional data space into a lower-dimensional feature space and define centroid-based clustering loss such as K-means and K-median~\cite{dasgupta2020explainable} on top of the embedded layer~\cite{xie2016unsupervised,guo2017improved,rezaei2021learning}. However, DEC is not able to model the generative process of data.

We propose a novel deep generative variational autoencoder framework for simultaneously learning unsupervised representation and perform image clustering. Our proposed DVC is able to model data generative process by multivariate Gaussian model and deep convolutional neural network.
In order to perform deep embedded clustering on large-scale medical images, we develop the pipeline with a deep convolutional neural networks which results in a better quality of feature maps, reducing the number of parameters, and preserving locality since weights are shared among all locations in the input. 

In summary, the convolutional autoencoder is utilized to learn representations in an unsupervised way where the learned features can preserve essential local structure in data. Our probabilistic encoder utilizes the multivariate Gaussian and captures the latent distribution of a large set of unlabeled images. The clustering loss is applied on top of encoder features and helps to scatter embedded points. However, training with only clustering loss causes the corruption of latent space and results in inaccurate performance. Therefore, we propose a probabilistic decoder to prevent the corruption of data points in the latent space. The proposed decoder network modifies embedded space and helps to separate the clusters accurately.

\section{Related works and Background} \label{related}
\noindent\textbf{Deep Clustering} In recent years, several approaches perform clustering on top of features extracted by deep neural network (DNN) ~\cite{constrainedclustering2019,cao2019learning,rezaei2021learning}. Tian et al.~\cite{tian2014learning} proposed a two-stage framework that runs K-means clustering on the feature space extracted by a DNN in the first stage. The proposed framework by Peng et al.~\cite{peng2016deep} includes a sparse autoencoder that learns representations in nonlinear latent space, followed by conventional clustering algorithms to fulfill label assignment. 

Deep embedded clustering~\cite{xie2016unsupervised} trains an autoencoder with a reconstruction loss paired with a cluster assignment loss. It then defines a soft cluster assignment distribution by using k-means on top of the learned latent representations. The algorithm was later improved by an additional reconstruction loss ~\cite{guo2017improved}(to preserve local structure), an adversarial loss~\cite{mrabah2020adversarial}, and data augmentation~\cite{guo2018deep}. JULE~\cite{yang2016joint} is an end-to-end deep clustering framework that jointly learns Convnet features and clusters within a recurrent framework. Bise et al.~\cite{constrainedclustering2019} proposed a soft-constrained clustering method on top of CNN's features and applied it for clustering of endoscopy images.

\noindent\textbf{Deep Generative Clustering} 
Deep generative models are the powerful class of machine learning which are able to capture the data distribution of the training data and generate artificial samples. This makes it suitable for bioinformatics use cases with limited labeled or unlabeled samples~\cite{rezaei2020generative,rezaei2020generativemodel,rezaei2020generativecardiac}. Generative adversarial networks (GANs) and variational autoencoders (VAEs) are the most popular and efficient approaches among generative models. In the context of image clustering, ClusterGAN~\cite{mukherjee2019clustergan} performs clustering with GAN framework and additional deep clustering network which trained with three players in an adversarial fashion. 

Jiang et al.~\cite{jiang2016variational} introduce variational deep embedding (VaDE) and use Gaussian Mixture and VAE together for building the inference model. Similarly, DGG~\cite{yang2019deep} and GMVAE~\cite{dilokthanakul2016deep} exploit VAE and GMM to minimize the graph distances between embedding data points. The mentioned methods and also other previous studies such as ~\cite{jiang2016variational,yang2019deep,mukherjee2019clustergan} do not address problems arising due to local preservation. Here, DVC can simultaneously optimize cluster labels assignment and learn features that are suitable for clustering with local structure preservation by combining the clustering loss, KL loss, and autoencoder’s reconstruction loss.

\noindent\textbf{Variational Autoencoder} 
VAEs are a probabilistic twist of autoencoders that approximate data distributions by optimizing evidence lower bound loss~(ELBO)~\cite{kingma2013auto}: 

\begin{equation} \label{eq_vae}
\mathcal{L}_{n}(\theta,\phi;x,z) = -\mathbb{E}_{q_{\phi} (z|x)} [\log p_\theta (x|z)] + \mathcal{D}_{KL} (q_\phi (z|x) \Vert p(z)) 
\end{equation}
where the first term is a reconstruction loss and the second is a regularization term, $\phi$ and $\theta$ are the parameters of encoder and decoder network respectively. The $p(z)$ is a fixed prior distribution on latent distribution with a common choice of normal Gaussian:

\begin{equation}
p(z) = \mathcal{N} (\mu=0, \sigma^2=I)
\end{equation}

The $q_{\phi}(z|x)$ and $p_{\theta}(x|z)$ in Eq.(\ref{eq_vae}) are diagonal normal distributions parameterized by neural networks can be computed from:

\begin{equation} \label{eq.1}
q_{\phi}(z|x) = \mathcal{N}(z;\mu,\sigma^2*I)
\end{equation}

\begin{equation} \label{eq.2}
p_{\theta}(x|z) = \mathcal{N} (x;\mu, \sigma^2 * I) \quad \textrm{or} \quad Ber(x; p_\theta(z))
\end{equation}
However, VAEs performance influenced by the design of network architecture and choosing hyperparameters such as the size of latent variables, input and output dimension, and standard deviation of $p(x|z)$. Considering a pretrained model, sufficiently powerful neural networks, a large enough latent space, VAEs with Gaussian encoders and decoders can approximate the true data distribution. Therefore, after optimization, $\mathcal{L_{ELBO}}$ is often used as a proxy for the likelihood of a data sample.

\section{Method} \label{method}

\begin{figure*}[!t]
\includegraphics[width=0.93\textwidth]{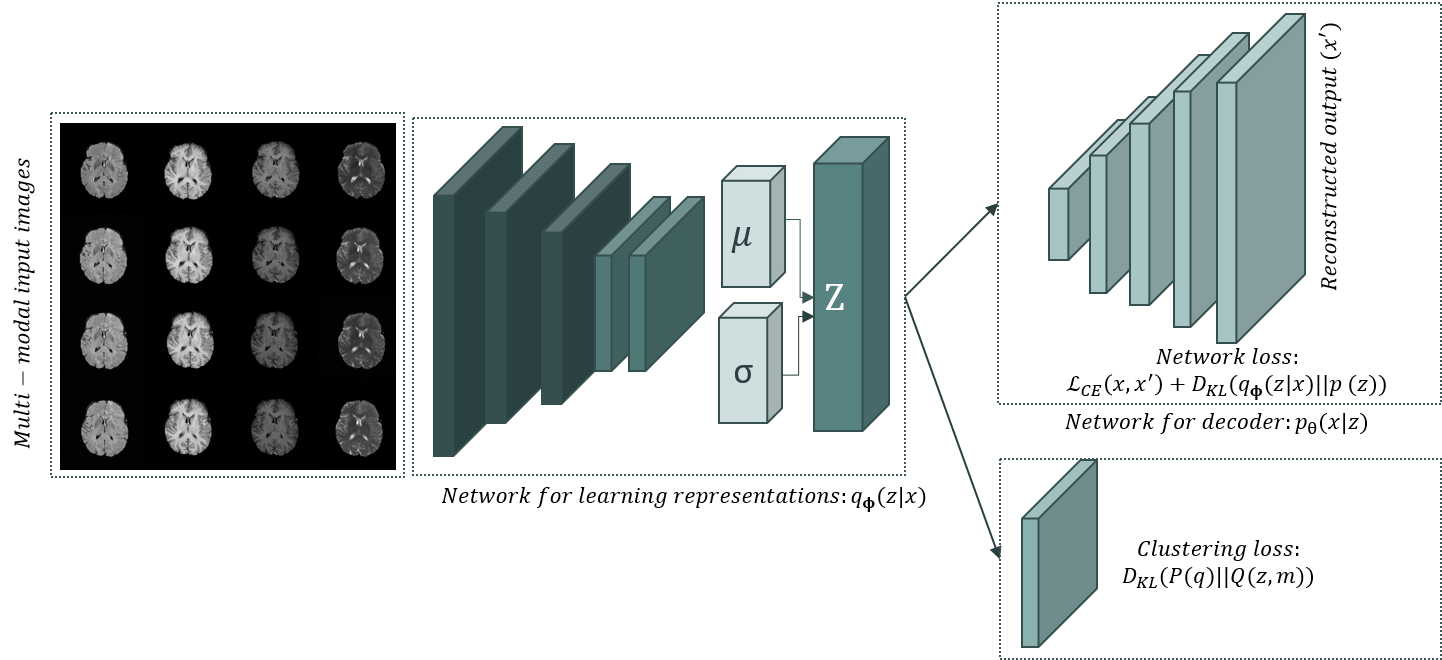}
\caption{Illustration of the proposed deep variation clustering network. The large-scale clustering is performed on top of deep features extracted by the convolutional neural network, variational embedded layer, and by minimizing the \textit{KL} divergence loss between a prior distribution and extracted features by the encoder network. 
} \label{fig1}
\end{figure*}
 

Our goal is to cluster $N$ samples $\{x_i\}_{i=1}^N$ from the input space $\mathcal{X}=\mathbb{R}^{d_x}, d_x \in \mathbb{N}$ into $K$ clusters, represented by centroids $m_1,\ldots,m_K \in \mathbb{R}^{d_x}$. The proposed framework is composed of two networks (see Fig \ref{fig1}). The encoder network $q$ with parameters of $\phi$ computes $q_\phi (z|x): x_i\rightarrow z_i$. The encoder maps an input image $x_i \in \mathcal{X}$ to its latent embedding $z_i\in\mathcal{Z}$ in a lower dimensionality space compared to the input space $\mathcal{X}$. The decoder network $p$ parametrizes by $\theta$, $p_{\theta}:z_i\rightarrow x_i^\prime$, and reconstructs $x_i$ from its latent embedding $z_i$. The autoencoder network is first pre-trained, with the network loss $\mathcal{L}_n = \mathcal{L}_r + D_{KL}(q_{\phi}(z|x) \Vert p(z))$, where $\mathcal{L}_r = \mathcal{L}_{CE}(x, x')$, to initialize latent variable $z_i$. After convergence, the clustering loss $\mathcal{L}_c = D_{KL}(P(q)\Vert Q(z, m))$ is introduced into the objective function, and the network's training is continued to jointly perform clustering in the embedded space while preserving reconstruction capabilities. This is done by using a convex combination of the two losses. The relative weight of each of the two losses is indicated by $\lambda \in (0,1)$, which controls the degree of distortion introduced in the embedded space:

\begin{equation} \label{eq_1}
\mathcal{L} =  \lambda \mathcal{L}_{c} + (1-\lambda) \mathcal{L}_{r}.
\end{equation}



In other word, the training procedure of the proposed DVC is end-to-end in two-phase; In the first phase, we initialize the VAE parameters with a multivariate Gaussian samples. At the second phase, the parameters are optimized by training simultaneously deep variational embedded clustering and deep probabilistic reconstruction network. Therefore, we iterate between: computing an auxiliary target distribution and minimizing KL divergence to the computed auxiliary target distribution.

\subsection{Deep Variational Clustering and Parameter Initialization} \label{method-init}
After convergence of the first training step of our network, yielding a good embedding representation of each training sample in the first step, our method performs clustering in the latent space $\mathcal{Z}$. We use the Kullback-Leibler (KL) divergence as clustering loss:

\begin{equation} \label{eq_3}
\mathcal{L}_c=\text{KL}(P\Vert Q) = \sum_{i} \sum _{j} p_{ij}\ln {\frac {p_{ij}}{q_{ij}}}
\end{equation}

\noindent where $Q$ is a soft labeling distribution with elements $q_{ij}$. $P$ is an auxiliary target distribution derived from $Q$ with elements $p_{ij}$. More specifically, $p_{ij}$ are the elements of the target distribution while $q_{ij}$ is the distance between the embedded $z_i$ and the center $m_j$ of the $j$-th cluster. This distance is measured by a Student's \textit{t-}distribution (cf. ~\cite{van2009learning}):

\begin{equation} \label{eq_4}
q_{ij} = \frac{(1+\norm{z_i - m_j}^2 / \alpha)^{-\frac{\alpha +1}{2}}}{\sum_{k}(1+ \norm{z_i - m_k}^2 / \alpha )^{-\frac{\alpha +1}{2}}}
\end{equation}

\noindent where $\alpha$ is the degrees of freedom of the Student’s \textit{t-}distribution (we here only consider $\alpha=1$).

In order to address problems associated with small disjuncts, we modify the target distribution $P$ by pushing data points that are similar in the original space closer together in the latent space. Thereby samples from less-frequent classes can be identified as a cluster. Then, $p_{ij}$ is computed as follows:

\begin{equation} \label{eq_5}
p_{ij} = \frac{q_{ij}^2 / (u_{j} + v_{j} )}{\sum_k q_{ik}^2 / v_{k}}
\end{equation}

\noindent where $u_{j} = \sum_i q_{ij}$ are soft cluster frequencies while $v_{j}$ normalize the frequency of samples per cluster:

\begin{equation} \label{eq_5_2}
v_{j} = - \sum_i \sum_j \sqrt{ \frac{ \sum_k N_k}{N_j} {(1-q_{ij})}^\gamma \log (q_{ij})}.
\end{equation} 

\noindent Here, $N_j$ is the estimated cardinality of cluster $j$, $\gamma$ is a relaxation parameter in laymen’s terms and set to 2. Note that, to prevent instability in the training procedure, we do not update $P$ at every iteration. $P$ is only updated if changes in the label assignments between two consecutive updates of the target distribution are less than a threshold $\delta$. This tolerance threshold and its empirical property are discussed in more detail in Section~\ref{discussion}.

For parameter initialization, we follow the standard procedure by \cite{xie2016unsupervised} and \cite{guo2017improved}: the autoencoder is pre-trained separately, and the centroids $m_1,\ldots,m_K$ are initialized by performing standard $K$-means clustering on the latent embeddings of the training samples.

Both probabilistic encoder and decoder build by convolutional neural network architecture which brings the following advantages: 1) suitable for large-scale medical images, 2) better quality of feature maps in which results in better representation 3) fewer parameters and hyperparameter needed to be tuned.

\noindent\textbf{Self-labeling and Optimization} We perform multi-objective optimization to jointly optimize the cluster loss and the reconstruction loss (ELBO) using mini-batch stochastic gradient variational Bayes (SGVB). In each iteration, the probabilistic encoder network’s weights~$\phi$, cluster centers, probabilistic decoder's weights~${\theta}$, and target distribution~$P$ are updated and optimized on the basis of \eqref{eq_1}.
The target distribution $P$ plays as the ground-truth of the soft label. By iterating these updates, the label assigned of $x$ is obtained using:

\begin{equation} \label{eq_15}
y_i =   \underset{j}{\mbox{argmax}} \,  q_{ij},
\end{equation}

\noindent The overall training process is repeated until a convergence criterion based on the KL loss is met. Algorithm~\ref{alg1} summarizes the training procedure.

\begin{algorithm}[!t]
\SetAlgoLined
    \SetKwInOut{Input}{Input}
    \SetKwInOut{Output}{Output}
    \Input{input data 
    $\mathcal{X}=\mathbb{R}^{d_x}$, initial number of clusters $K$, convergence threshold $\delta$}
    \Output{cluster label $y_i$ of $x_i \in \mathcal{X}$}
     initialize $\theta$, ${\phi}$ as described in~\ref{method}\;
     initialize $m_1,\ldots,m_K$ using K-means\;
     \For{$itr \in \{0, 1,$\ldots$, Itr_{max}\}$ }{
         \eIf{not converged}{
           \eIf{itr\%update\_interval==0}{
            calculate all embedded points $\{z_i = f_{\theta}(x_i)\}^n_{i=1}$\;
            calculate student's t-distribution 
            $Q, q_{ik} = \frac{(1 + \norm{z_i - m_k}^2)^{-1}}{\sum_k (1 + \norm{z_i - m_k}^2)^{-1}}$ (Eq.~\ref{eq_4})\;
            update target distribution $P, p_{ij} = \frac{q_{ij}^2 / (u_{j} + v_{j} )}{\sum_k q_{ik}^2 / v_{k}}$ (Eq.~\ref{eq_5})\; 
            do soft label assignment (Eq.~\ref{eq_15}) \;
            }{
            select a mini-batch of samples \;
            calculate $z_i$ and $q_i$ for each mini-batch\;
            calculate $x_i^\prime = f_{\phi}(z_i)$  \;
            compute $\mathcal{L}_r$ and ${L}_c$\;
            update $m_1,\ldots,m_K, {\theta}, {\phi}$\;
            }
        }{Stop training.}
    }
 \caption{Deep Variational Clustering Algorithm}
 \label{alg1}
\end{algorithm}

\section{Experiments} \label{Experiments}
In this section, we conduct experiments to examine our proposed framework. First, we compare the results of our method with several of the related state-of-the-art methods on various clustering task.
\\
\\
\noindent\textbf{Datasets} The proposed method is evaluated on MNIST~\cite{lecun1998gradient}, BRATS 2018~\cite{Menze2014,Bakasnature2017,Bakastcg2017,Bakaslgg2017} and REFUGE-2~\cite{orlando2020refuge} image datasets.
\\
\noindent\textit{MNIST} consists of 60,000 images for training and 10,000 for testing, each image has a size of $28 \times 28$ pixels and is from one of 10 classes. We train on the full training set and report as well as compare the results to other methods on the test set.
\noindent \textit{BRATS 2018} \cite{Menze2014,Bakasnature2017,Bakastcg2017,Bakaslgg2017}
consists of two different brain diseases; high and low-grade glioma (HGG/LGG) brain tumour(s). All brains in the dataset have the same orientation. The dataset composed of 1156 magnetic resonance images in four modalities T2, Flair, T1, and T1c from 724 HGG patients and 432 LGG patients.
\\
\noindent \textit{REFUGE-2}~\cite{orlando2020refuge} is an active challenge on classification of clinical Glaucoma and part of the MICCAI-2020 conference. The organizers released 1,200 microscopy retinal scans with size of $2124 \times 2056$ pixels from two different machines and scanned by two clinics. The dataset is imbalanced with a ratio of 1:30.  
\\
\\
\noindent\textbf{Evaluation Metrics}
As unsupervised evaluation metric, we use the clustering Accuracy (ACC), Normalized Mutual Information (NMI) and Adjusted Rand Index (ARI) for evaluations. These measures range in $[0, 1]$, higher scores show more accurate clustering results.

\noindent\textbf{Compared Method}
We compare and show the efficacy of our proposed algorithm by comparing with unsupervised DEC~\cite{xie2016unsupervised}, IDEC~\cite{guo2017improved}, DEPICT~\cite{ghasedi2017deep}, VaDE~\cite{jiang2016variational}, variational clustering~\cite{prasad2020variational}, and VDEC~\cite{ghosh2019variational}, which can be viewed as a variant of our method when the reconstruction loss and network architecture are different. Note that the reported results for DEC and IDEC is based on our implementation code in pytorch.
\\
\\
\noindent\textbf{Experimental Setting} ~\label{discussion}
We evaluate two architectural choices for our proposed framework by modifying different types of auto-encoders. In the first experiment (\emph{DVC-1}), we study the impact of convolutional neural network architecture and combination of \textit{KL}-loss and binary cross-entropy loss for large scale image clustering. Both probabilistic encoder and decoder build by convolutional neural network architecture. We use convolutional layers with kernel size 5 $\times$ 5 and stride 2 in the encoder part. In the decoder part, we perform up-sampling by image re-size layers with a factor of 2 and a convolutional layer with kernel size 3 $\times$ 3 and stride 1. The input size for images from REFUGE and BRATS is 128 $\times$ 128. Using convolutional encoder and decoder are beneficial for medical images which are usually large-scale and high-dimensional, and also can bring feature maps and representations. Moreover, with convolutional layers, fewer parameters and hyperparameter are required to be considered.

\begin{figure*}[!t]
\includegraphics[width=\textwidth]{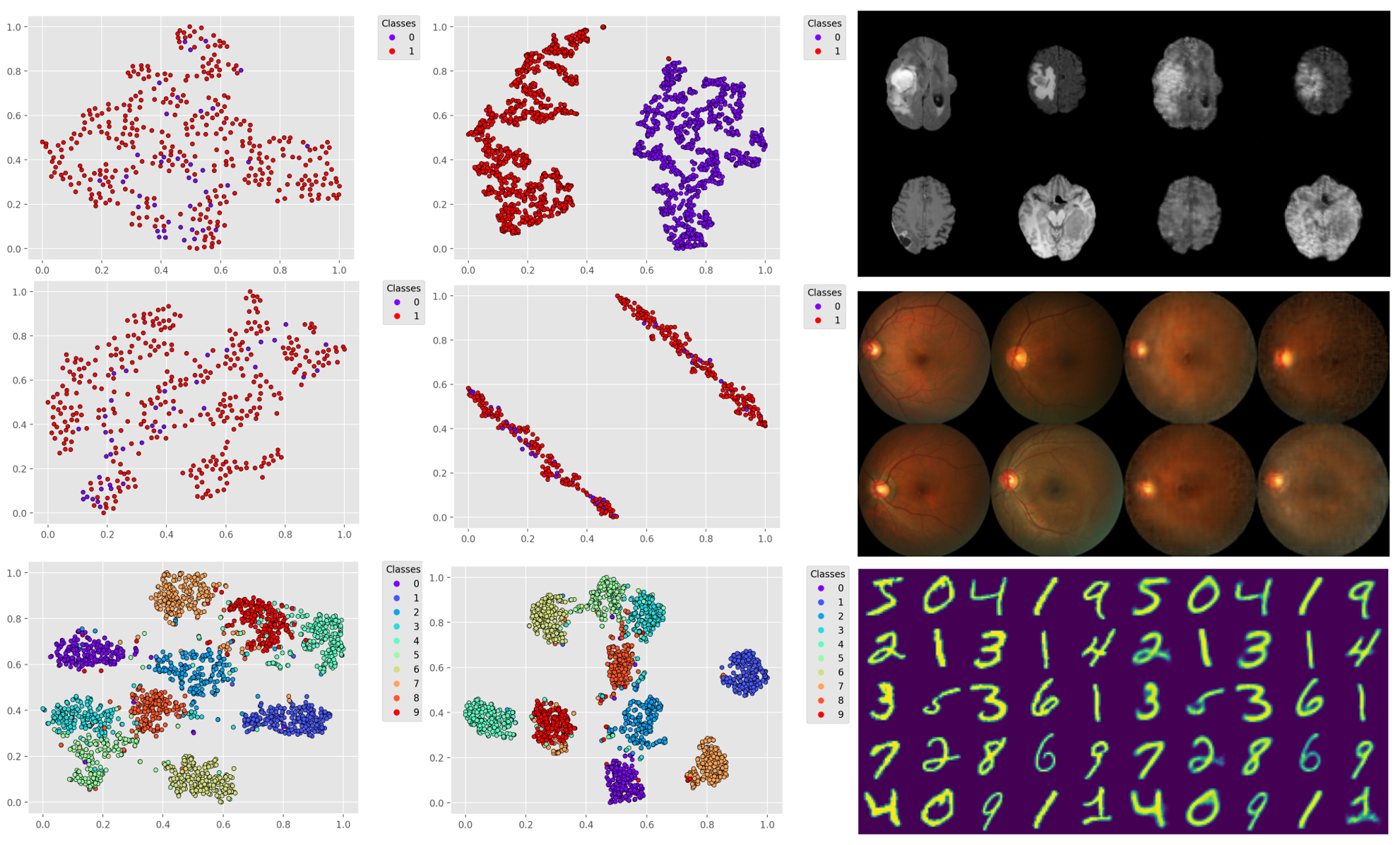}
\caption{Illustration of the qualitative results achieved by our proposed method. Left) the clusters in the first epoch, Middle) computed clusters in the final epoch, Right) Several samples generated by our trained model. Top to bottom is BRATS, REFUGE, and MNIST dataset. The results show the promising functionality of the proposed method.} \label{fig2}
\end{figure*}

The second experiment \emph{DVC-2}, includes a fully-connected multilayer perceptron (MLP) with dimensions \textit{$d_x$-500-500-1000-10} as encoder for all datasets. Here, $d_x$ is the dimension of the input data. The decoder network is also a fully-connected MLP with dimensions \textit{10-1000-500-500-$d_x$}. Each layer is pre-trained for 100,000 iterations with dropout. The entire deep autoencoder is further fine-tuned for 200,000 iterations without dropout for both layer-wise pre-training and end-to-end tuning. The mini-batch size is set to 256 for MNIST, 64 for BRATS , and 8 for REFUGE. We use a learning rate of 0.01 which is divided by 10 every 20,000 iterations and set weight decay to zero. After pretraining, the coefficient $\lambda$ of clustering loss is set to 0.1. The convergence threshold $\delta$ is set to 0.001 while the update intervals for target distribution are 70, 80, 100 iterations for REFUGE, BRATS, and MNIST respectively.
\\
\\
\noindent\textbf{Image Clustering} 
We include the quantitative clustering results of these clustering methods in Table~\ref{table1}. Based on columns 4-9, our proposed DVC-1 outperforms the other methods with significant margins on all three clustering quality measures. Other points that can be also observed from Table~\ref{table1} are: (1) The performance of convolutional-based methods (DVC-1) is much better than those fully connected MLP methods (DVC-2 and VAE~\cite{kingma2013auto}). The reason is the quality of the feature map and fewer parameters that need to be tuned. (2) The performance of representation-based clustering algorithms (DAC~\cite{chang2017deep}) is much better than the conventional clustering algorithms (i.e. K-means~\cite{wang2014optimized}). It shows that representation learning acts a crucial role in image clustering. The qualitative results are shown in Figure \ref{fig2} includes the cluster distributions and several generated image samples. Figure \ref{fig2} shows promising abilities of the proposed method.

\begin{table*} [!htbp]
\centering
\caption{Comparison results of our achieved accuracy on MNIST, BRATS and REFUGE datasets. Results of (*) methods are taken from reference ~\cite{chang2017deep}.}
\label{table1}
\begin{tabular}{l l l l |l l l |l l l}
\toprule
Model  & \multicolumn{3}{c}{MNIST~\cite{lecun1998gradient}} &  \multicolumn{3}{c}{REFUGE~\cite{krizhevsky2009learning}} &
\multicolumn{3}{c}{BRATS~\cite{Menze2014,Bakasnature2017,Bakastcg2017,Bakaslgg2017}} \\
\midrule
  & NMI & ARI & ACC & NMI & ARI & ACC & NMI & ARI & ACC  \\
\midrule
DVC-1 & 0.798 & 0.777 & 0.888 & 0.712 & 0.697 & 0.751 & 0.943 & 0.951 & 0.963 \\
DVC-2 & 0.767 & 0.786 & 0.876 & 0.653 & 0.583 & 0.716 & 0.889 & 0.892 & 0.901 \\
\midrule
*K-means \cite{wang2014optimized} & 0.499 & 0.365 & 0.572 & 0.083 & 0.028 & 0.129 & 0.556 & 0.583 & 0.612\\
AE~\cite{bengio2006greedy} &  0.725 & 0.613 & 0.812 & 0.100 & 0.047 & 0.164 & 0.582 & 0.607 & 0.679 \\
VAE~\cite{kingma2013auto} & 0.736 & 0.712 & 0.831 & 0.407 & 0.440 & 0.521 & 0.604 & 0.631 & 0.682 \\
*JULE~\cite{yang2016joint} & 0.913 & 0.927 & 0.964 & 0.602 & 0.632 & 0.662 & -  & - & - \\
IDEC ~\cite{guo2017improved}& 0.816 & 0.868 & 0.880 & 0.627 & 0.642 & 0.681 & 0.837 & 0.844 & 0.892 \\
DEC~\cite{xie2016unsupervised}& 0.771 & 0.741 & 0.843 & 0.513 & 0.549 & 0.585 & 0.805 & 0.842 & 0.853\\
VDEC~\cite{ghosh2019variational} & 0.836 & 0.748 & 0.842 & 0.612 & 0.645 & 0.679 & 0.833 & 0.874 & 0.901  \\
*DAC~\cite{chang2017deep} & 0.935 & 0.948 & 0.975 & 0.703 & 0.717 & 0.734 & - & - & - \\
\bottomrule
\end{tabular}
\end{table*}

\section{Conclusion} \label{conclusion}

We proposed end-to-end CNN-based VAEs with multivariate Gaussian priors to perform unsupervised image clustering. We performed clustering on top of strong latent representation made with both prior and posterior distribution. In comparison with the existing approaches, the proposed method achieves superior performance on two real patient medical imaging datasets and competitive results on the MNIST and CIFAR-10 datasets. It shows that our method can deal with large-scale medical images in different image modalities and is not limited to simple image datasets. We presented the application of our proposed method for the task of unsupervised clustering. However, the synthesized samples look realistic with high-resolution therefore they can tackle some DL challenges such as class imbalance and data augmentation. 

\section{Acknowledgment}
This work has been funded in part by the German Federal Ministry of Education and Research (BMBF) under Grant No. 01IS18036A, Munich Center for Machine Learning (MCML). F.S, M.R, and B.B were supported by the German Federal Ministry of Education and Research (BMBF) under Grant No. 01IS18036A. The authors of this work take full responsibilities for its content.

\bibliographystyle{IEEEtran}
\bibliography{icdm}

\end{document}